\documentclass{article}

\PassOptionsToPackage{numbers, compress}{natbib}

\usepackage[final]{nips_2017}

\usepackage[utf8]{inputenc} 
\usepackage[T1]{fontenc}    
\usepackage{hyperref}       
\usepackage{url}            
\usepackage{booktabs}       
\usepackage{amsfonts}       
\usepackage{nicefrac}       
\usepackage{microtype}      

\usepackage{graphicx}

\title{Relationships from Entity Stream}

\author{
  Martin Andrews\\
  Red Dragon AI\\
  Singapore\\
  \texttt{martin@reddragon.ai} \\
  \And
  Sam Witteveen\\
  Red Dragon AI\\
  Singapore\\
  \texttt{sam@reddragon.ai} \\
}

\begin{document}

\maketitle

\begin{abstract}
  Relational reasoning is a central component of intelligent behavior, 
  but has proven difficult for neural networks to learn. 
  The Relation Network (RN) module was recently proposed by DeepMind to solve such problems, 
  and demonstrated state-of-the-art results on a number of datasets. 
  However, the RN module scales quadratically in the size of the input, 
  since it calculates relationship factors between every patch in the visual field, 
  including those that do not correspond to entities.
  In this paper, we describe an architecture that enables relationships to be determined 
  from a stream of entities obtained by an attention mechanism over the input field.
  The model is trained end-to-end, and demonstrates 
  equivalent performance with greater interpretability 
  while requiring only a fraction of the model parameters of the original RN module.  
\end{abstract}

\section{Introduction}



The ability to reason about the relations between entities and their properties 
is central to intelligent behavior (\citet{kemp2008discovery,johnson2016clevr}). 




Our work shows how a deep learning architecture equipped 
with attention, sequencing and selection mechanisms
can learn to identify relevant entities in visual field, and then reason
about the relationships between them as part of a stream.

One appealing feature of this approach is that the Entity Stream it creates 
can be thought of as a grounded (\citet{harnad1990symbol}) internal representation 
that is then manipulated on a more symbolic level in order to answer questions about the scene.

\section{Related Work}

Recently, \citet{DBLP:journals/corr/SantoroRBMPBL17} highlighted 
the problem of grounding visual symbols so that relationships between
the entities could be determined.  However, the Relation Network technique 
introduced there requires that every potential entity in the visual scene 
to be `related' to every other entity.  
Moreover, even blank spaces were treated as entities : 
No actual grounding is being done.

As also pointed out in \citet{DBLP:journals/corr/abs-1709-07871} 
(which established new state-of-the-art results on the CLEVR dataset 
using repeated applications of affine transformations to focus on 
parts of the source image), the RN method is quadratic in 
spatial resolution, which, in part, was a key motivation for the present work.

The approach of \citet{DBLP:journals/corr/HuARDS17} is closer to treating 
entities symbolically, by creating an environment that includes 
primitives such as searching, sorting and filtering 
(i.e. much `heavier machinery' than used here). 
In \citet{DBLP:journals/corr/BansalNM17}, entities are stored in a collection 
of memory elements, and relationships are sought amongst them there.


\section{Relationships from Entity Streams}

\begin{figure}[hb]
  \centering
  \includegraphics[width=1.0\textwidth]{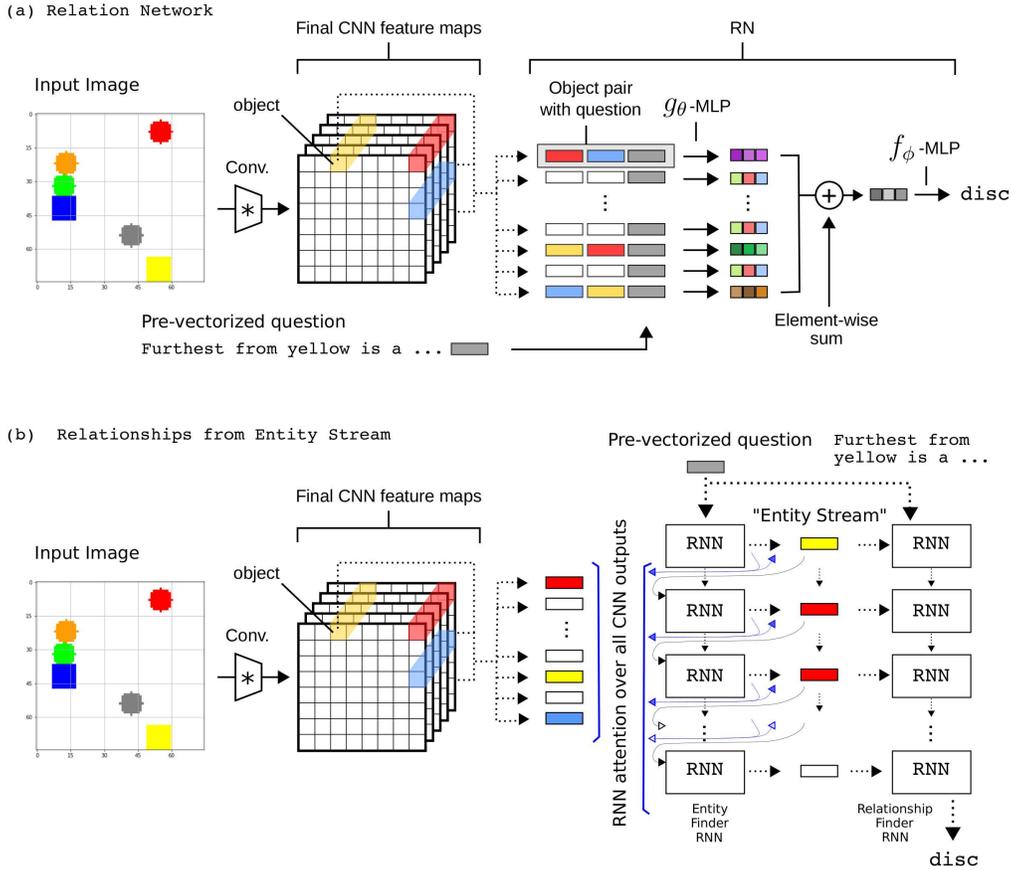}
    \caption{
      \textbf{Model architectures}. 
      (a) The original Relation Network diagram, updated to show how all patches in the 
      input image are being compared to each other;  
      (b) Relationships from Entity Stream, in which the same preprocessing occurs, 
      but the entities are selected through an attention mechanism by the left-hand column of recurrent units, 
      resulting in an "Entity Stream".  
      The right-hand column of recurrent units then interprets the Entity Stream to answer the question.
    }
    \label{fig:architecture}
\end{figure}

\subsection{Architecture}

In order to evaluate architectures on an equal footing, the image preprocessing
layers were the same as in \cite{DBLP:journals/corr/SantoroRBMPBL17}, where
the Sort-of-CLEVR image passes through CNN layers to create a $5\times 5$ field of patches,
each position of which obtains a $24$ dimensional vector as a representation.

This vector field is then split arbitrarily into \emph{key} and \emph{value} portions, 
of length $14$ and $10$ respectively.
Similar to the RN paper, coordinate-related elements are also appended, so that the model
can reason about geometry in later stages.
As shown in diagram (b) of Figure~\ref{fig:architecture}, the data passes through two different RNN columns.  
Both of these (the `entity finder' EF, and the `relationship finder' RF) consist of two-layer stacked GRUs, and
their initial hidden states are fed with the question vector (which is of length $11$), padded with trainable parameters
to the GRU hidden dimension size.

The output hidden state of the EF RNN is used as the \emph{query} for attention over the patches, 
as described in \cite{DBLP:journals/corr/VaswaniSPUJGKP17}.
The per-patch weights then undergo a $SoftMax$ operation (with optional hardening, see below) 
to then weight the patch \emph{values}.
The attention-weighted sequence of \emph{values} retrieved by the EF RNN 
is (reasonably) denoted the "Entity Stream".  
The entity value found at a given timestep is also used as input 
to the EF RNN on the next timestep, so that it can produce 
a coherent sequence of values for the Entity Stream.

The Entity Stream is then used as the input sequence for the RF RNN 
and the model is trained end-to-end given the input image, vectorized questions, and vectorized answers 
(which are matched using categorical-cross-entropy against the RF RNN's final hidden state).

\subsection{Soft vs Hard attention}

The proposed architecture allows for flexibility in construction of the `entities' 
for the Entity Stream, which are the \emph{value} portion of the CNN output from the visual scene, 
and so will typically be a mix of colour, shape and location information.

For the RFS model, the weights derived from the attention mechanism are simply passed 
through a $SoftMax$ function, so that it is possible for the Entity Stream to 
contain `entities' that are averages of the information from several patches in the image.
While simple to implement, this behaviour enables the network to `cheat' for counting questions, 
since it can (for instance) simply attend to all square objects, 
and measure the variety of the response in order to count in one operation/timestep.

To eliminate this effect, we also propose the RFSH (the Hard attention version of RFS), 
where the weights from the attention mechanism have Gumbel random variables added to them, 
and a hard $OneHot$ weight is used, as described in \citet{2016arXiv161101144J}.  
This causes the Entity Stream to be `pure', 
without having to train the model using REINFORCE-style learning.

\section{Experiments : Sort-of-CLEVR}



The experimental set-up used here is the Sort-of-CLEVR data, as first
described in \cite{DBLP:journals/corr/SantoroRBMPBL17}.
Examples of the input images, and question/answer pairs are illustrated in Figure~\ref{fig:attention}.

%

Because the dataset is visually simple, which reduces complexities involved in image processing, 
and the questions and answers have pre-defined vector representations, 
the Sort-of-CLEVR dataset is ideal for rapid experimentation, 
with the training / testing over 50-100 epochs typically taking less than 2 hours on a typical GPU.

The code for the 4 different models (and dataset generation) 
can be found at \url{https://github.com/mdda/relationships-from-entity-stream}

\section{Results}

A comparison of the final performance of the different methodologies is given in Table~\ref{table:perf}.
While the performance of the 3 purpose-built methods (RN, RFS and RFSH) are basically
equivalent, it is the qualitative differences that separate them.


\subsection{Attention Maps}

\begin{figure}[ht]
  \centering
  \includegraphics[width=0.95\textwidth]{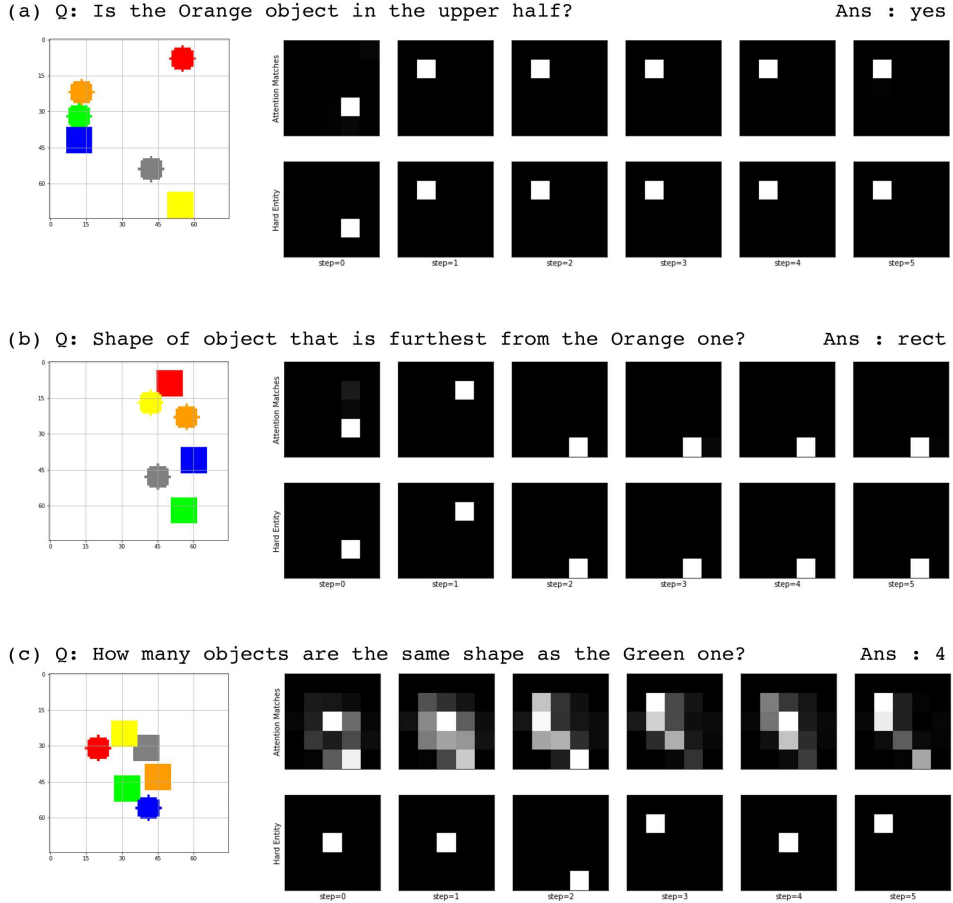}
    \caption{
      \textbf{Sample attention maps for RFSH model}. 
      For each of the 3 questions, the diagram shows the input image, as well as the attention
      match heatmap (the degree by which each patch in the image responds to the `entity finder' RNN at each timestep)
      in the upper row, and the `hard entity' chosen of the Entity Stream in the lower row,
      For cases (a) and (b), the model has learned to simply match the entities that it needs 
      (so the match heatmap is almost one-hot already).  
      As there is no benefit for getting early results, the model seems to sample the first timestep `carelessly'.
      For case (c), the question involves counting, and the entities are closely spaced, 
      so the model is more clearly conflicted about which to pick.  
    }
    \label{fig:attention}
\end{figure}

As illustrated in Figure~\ref{fig:attention}, the RFSH model learns to answer relationship
questions about shapes in the input image by creating a useful attention model;
choosing elements to put into its Entity Stream; and then determining 
the answer to each question from the Entity Stream itself.

The attention map visualization indicates that the RFSH model has learned something 
about the structure of the problem.  
This is qualitatively different approach than the more standard methods 
of combining images features and a representation of the question
to map to an answer vector.

\begin{table}[hb]
  \caption{Network performance summary}
  \label{table:perf}
  \centering
  \begin{tabular}{lrrrr}
    \toprule
    Model    & \multicolumn{1}{p{2cm}}{\raggedleft NonRel \\ fraction correct}  &  \multicolumn{1}{p{2cm}}{\raggedleft BiRel \\ fraction correct} &  hidden\_dim  &   Size in bytes \\
    \midrule
    RN      & 99\%  & 93\%  &  - &  1,463,513 \\
    CNN     & 98\%  & 63\%  &  - &    970,874 \\
    RFS     & 99\%  & 95\%  & 32 &    166,380 \\
    RFSH    & 99\%  & 93\%  & 64 &    408,364 \\
    \bottomrule
  \end{tabular}
\end{table}

\section{Further Work}


It would be interesting to explore whether several Entity Streams
could be used compositionally - since that is where the 
Relationships from Entity Stream approach
clearly differentiates itself from methods that are more tied
to the structure of the input space.  

One other interesting line of enquiry is to use sampling methods 
like Self-critical Sequence Training \cite{DBLP:journals/corr/RennieMMRG16} 
to learn more difficult cases (where the rewards are both delayed, 
and non-differentiable).  

\section*{Acknowledgements}

The authors are proponents of research results being verifiable through 
the release of working code.  Fortunately, the RN paper was quickly followed up 
with a release of an independent open source implementation by Kim Heecheol, 
on GitHub : \url{kimhc6028/relational-networks}.


\newpage
\bibliographystyle{plainnat}
\bibliography{relationships-from-stream_mdda}

\end{document}